\documentclass[11pt]{article}

\usepackage{graphicx}
\usepackage{booktabs}
\usepackage{multirow}
\usepackage{float}
\usepackage{hyperref}
\usepackage{amsmath}
\usepackage{amssymb}
\usepackage{cite}

\title{Enhancing the Detection of Coronary Artery Disease Using Machine Learning}

\author{
Karan Kumar Singh\\
Department of Computer Science and Engineering\\
Sharda University, Greater Noida, India\\
\texttt{karankumarsingh7870@gmail.com}
\and
Nikita Gajbhiye\\
Department of Computer Science and Engineering\\
Sharda University, Greater Noida, India\\
\texttt{nikitagajbhiye.ng@gmail.com}
\and
Gouri Sankar Mishra\\
Department of Computer Science and Engineering\\
Sharda University, Greater Noida, India\\
\texttt{gourisankar.mishra@sharada.ac.in}
}

\begin{document}

\maketitle

\begin{abstract}

Coronary Artery Disease (CAD) remains a leading cause of morbidity and mortality worldwide. Early detection is critical to recover patient outcomes and decrease healthcare costs. In recent years, machine learning (ML) advance- ments have shown significant potential in enhancing the accuracy of CAD diag- nosis. This study investigates the application of ML algorithms to improve the detection of CAD by analyzing patient data, including clinical features, imaging, and biomarker profiles. Bi-directional Long Short-Term Memory (Bi-LSTM), Gated Recurrent Units (GRU), and a hybrid of Bi-LSTM+GRU were trained on large datasets to predict the presence of CAD. Results demonstrated that these ML models outperformed traditional diagnostic methods in sensitivity and specificity, offering a robust tool for clinicians to make more informed decisions. The experimental results show that the hybrid model achieved an accuracy of 97.07\%. By integrating advanced data preprocessing techniques and feature selection, this study ensures optimal learning and model performance, setting a benchmark for the application of ML in CAD diagnosis. The integration of ML into CAD detection presents a promising avenue for personalized healthcare and could play a pivotal role in the future of cardiovascular disease management.

\end{abstract}


\section{Introduction}

Coronary Artery Disease (CAD*) occurs when the coronary arteries, which carry oxygen and nutrients to the heart muscle, are constricted or obstructed. Other names for this ailment include coronary and ischemic heart disease [1-3]. It is now well-recognized as an ongoing disease that poses the greatest risk to human life [4]. As per a recent study, the United States scores highest for both the prevalence of heart disease and the percentage of people diagnosed with the illness [5]. Many people suffer shortness of breath, swollen feet, extreme tiredness, and other signs of heart disease. CAD is the most prevalent form of Cardio Vascular Disease (CVD) and a leading cause of chest pain, stroke, and heart attacks. Heart rhythm problems, heart failure, congenital heart defects, and CVD are all forms of heart illness [6]. As a consequence, detecting CAD is essential for modern civilization. One of the most useful tools for diagnosing and guiding therapy for CAD is coronary angiography (CAG), which evaluates luminal stenosis, plaque features, and disease activity [7].

In healthcare applications, Machine Learning (ML) helps improve the diagnosis process and is therefore extensively used in medical science. To understand complicated and nonlinear patterns related to the characteristics, ML algorithms minimize the error between the projected and actual outcomes by analyzing data [13]. Integrating ML into medical applications has greatly improved diagnostic processes. ML has recently found several uses, assisting with nearly every aspect of medical diagnosis. These include cancer, Parkinson's disease, thyroid illness, and several ocular ailments [14]. ML in the context of CAD diagnosis allows for the quick identification of CHD using Electrocardiograms (ECGs), Phonocardiograms (PCGs), Coronary Computed Tomography Angiography (CCTAs), and CAG [15]. A comprehensive overview of ML's use in CAD diagnosis is provided in this paper. The research objective of the paper follows:\\

•	To develop an ML model that can precisely predict the occurrence of CAD using patient data such as demographic, clinical, and laboratory parameters.\\

•	To compare the effectiveness of different ML algorithms (e.g., Decision Trees, Random Forest, Support Vector Machines, Neural Networks) for predicting CAD.\\

•	To explore feature selection methods that identify the most significant predictors of CAD from a comprehensive dataset, ensuring that irrelevant variables do not hinder the model's accuracy.\\

•	To enhance detection accuracy by incorporating advanced data preprocessing techniques, such as handling missing data, normalization, and outlier detection, to clean and optimize the dataset for better learning outcomes.\\

To validate the performance of the ML models using metrics such as Sensitivity, Specificity, Accuracy and area under the ROC curve (AUC) on training and unseen test datasets.

Machine learning techniques have been successfully applied in various healthcare and agricultural applications, including depression detection, kidney disease diagnosis, and plant disease classification, demonstrating their effectiveness in predictive analytics \cite{gajbhiye2025rait,gajbhiye2025otcon,singh2025icdlair,gajbhiye2025crop}.

\section{Literature Review}

 In this section, the authors provide the previous work based on the detection of CAD using ML methods. All the results of previous work are evaluated in terms of Accuracy, Precision, Recall, Sensitivity , Specificity and F1-score. The performance metrics associated with the current approaches that are being evaluated are summarized in Table 1.

\begin{table}[H]
\centering
\caption{Summary of the outcomes achieved by current ML-based approaches}
\begin{tabular}{|p{3.5cm}|p{3.5cm}|p{3.5cm}|p{3.5cm}|}
\hline
\textbf{Author [Reference]} & \textbf{Data Collection (No. of patients)} & \textbf{Methodology} & \textbf{Outcomes} \\ 
\hline

Ma et al., [16] & Southwest China (300,000) & XG-Boost, RF, and LR & The AUC for the suggested model was 0.70. \\ 
\hline

Mijwil et al., [18] & UC Irvine ML repository (300) & MLP, RF, SVM, KNN, LR, NB, and DT & The MLP algorithm reached an accuracy of over 88\%. \\ 
\hline

Hammoud et al., [19] & 1190 & LR, SVM, KNN, RF, DT, NB, and GB & The RF algorithm achieved the highest accuracy rates of 89.50\% before tuning and 94.96\% after tuning. \\ 
\hline

Asif et al., [20] & Heart disease dataset & RF, GB, and XG-Boost & The ensemble model attained remarkable results in terms of Sensitivity (96.37\%), Accuracy (94.53\%), and Precision (98.37\%). \\ 
\hline

Lin et al., [21] & MIMIC-IV (5757) & Cat-Boost & The suggested model had the strongest predictive performance, with an AUC of 0.760 and a training set of 0.831. \\ 
\hline

Yilmaz et al., [22] & Heart Disease Dataset (11) & SVM and RF & The SVM model achieved Specificity of 0.844, Sensitivity of 0.971, NPV of 0.887, PPV of 0.976, and F1-score of 0.816. \\ 
\hline

Garavand et al., [23] & Z-Alizadeh Sani dataset (303) & MLP, SVM, LR, J48, RF, KNN, and NB & RF achieved the highest Accuracy (78.96\%), Sensitivity (93.86\%), Specificity (51.13\%), MCC (0.5192), and ROC (0.8375). \\ 
\hline

Huang et al., [24] & Affiliated Hospital of Traditional Chinese Medicine of Southwest Medical University (2049) & RF, KNN, RBFNN, KRR, and SVM & RF achieved the highest Accuracy (78.96\%), Sensitivity (93.86\%), Specificity (51.13\%), MCC (0.5192), and ROC (0.8375). \\ 
\hline

Devi et al., [25] & Framingham dataset (4238) & RF, DT, and SVM & The system achieved an accuracy of 88\% using recursive feature removal with the RF model. \\ 
\hline

\end{tabular}
\end{table}

\section{Methodology}

Several recent studies have explored the use of machine learning and deep learning techniques for predictive modeling in healthcare and related domains. Previous works have demonstrated the effectiveness of optimized machine learning frameworks and hybrid deep learning architectures in solving complex classification problems \cite{gajbhiye2025rait,gajbhiye2025otcon,singh2025icdlair,gajbhiye2025crop}.

CAD is an important cause of mortality globally, requiring early detection and accurate diagnosis to prevent severe complications. Conventional diagnostic techniques, such as angiography, though accurate, are invasive and costly. Non-invasive diagnostic methods based on clinical data and medical images can potentially improve early detection, but their accuracy remains a challenge. The goal of this project is to develop a machine learning-based system to enhance the detection of CAD. By integrating an advanced Bi-LSTM algorithm, this project aims to build a robust system for CAD detection. This system will leverage clinical and imaging datasets to improve diagnostic accuracy and assist healthcare professionals in making informed decisions, thereby reducing the need for invasive procedures.

\begin{figure}[H]
\centering
\includegraphics[width=0.8\textwidth]{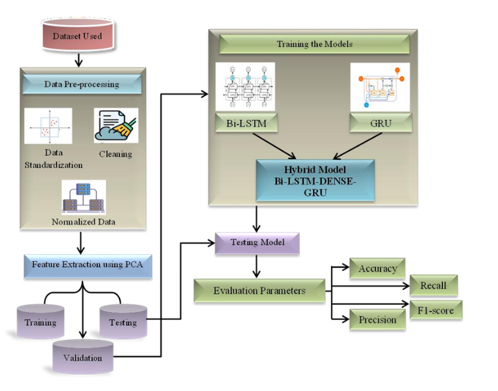}
\caption{Block diagram of proposed methodology}
\label{fig:methodology}
\end{figure}

\subsection{Data Collection}

A dataset of 1000 3D CCTA pictures has been created. A Siemens 128-slice dual- source scanner was used to collect the images. The images have a resolution of 512 × 512 × (206-275) voxels. The images were taken between 2012 and 2018 at the Shanti Hospital. On average, men were 67.68 years old, while females were 49.98 years old. Investigators can use the dataset repository [26] without restriction. Furthermore, it provides a way to segment images to get 3D views of coronary arteries.

\subsection{Data Preprocessing}

 Pre-processing is a fundamental step in any ML workflow. It involves preparing the dataset for training by transforming raw data into an understandable format for models. Pre-processing can include data cleaning, data standardization, and data normalization.
 
 Data Cleaning, sometimes known as data preparation, is an essential link in the data processing chain. The application of this method involves identifying and fixing any mistakes, inconsistencies, or missing values in the dataset. This is an essential step before doing any useful analysis of the dataset. In addition, the findings could be inflated, and the analysis might become confused if there are duplicate entries. This is achieved by removing the chance of accidental tampering, which helps ensure that the representational data is accurate [27].
 
Data Normalization is used to process the numeric data, which can avoid gradient dispersion, which is caused by the significant variance in individual features when employing the back-propagation technique [28]. A complicated dataset of features that DL struggles to extract is the consequence of not normalizing the data, which leads to a decline in gradient magnitude with back-propagation and a consequent slowing of the intrusion detection model's update weight increase. The collected CCTA dataset was normalized using the z-score method. to [-1, 1] using the z-score approach, as seen in Equation (1).

\begin{equation}
m'_i = \frac{m_i - \bar{m}}{x}
\label{eq1}
\end{equation}

The values of the data sample before and after normalization are denoted by mi and m'i, respectively. The average value of the feature's data before normalization is denoted by m.

Data Standardization was used to transform the dataset from a normal distribution into a standard normal distribution because it contains characteristics with varying ranges of values. Because of this, after rescaling, the mean value of an attribute is zero, and the standard deviation is the same as the distribution. The formula to compute a standard score (z-score) is:

\begin{equation}
z = \frac{x - \mu}{\sigma}
\label{eq2}
\end{equation}

The data sample, the mean, and the standard deviation are represented by $x$, $\mu$, and $\sigma$, respectively [29].

\subsection{Feature Extraction}

The strength and direction of the linear relationship between the two variables are measured by the coefficient of correlation, which can have values between -1 and 1. During the feature extraction process, the authors used the Pearson correlation matrix (PCM) [30] to determine which variables were most strongly connected. This allowed us to decrease the dataset's dimensionality while keeping all of the crucial information. In the end, this allowed us to analyze the data more efficiently and effectively. The PCM, which it uses to find the most associated variables, can be utilized as well to spot any abnormalities or outliers in the dataset [31]. Optional feature extraction techniques like PCA or Independent Component Analysis (ICA) could be more suited when the variables’ relationships are not linear.

\subsection{Classification Models}

Bidirectional Long Short-Term Memory (Bi-LSTM) is an enhancement to the traditional LSTM that has been suggested to improve the performance of network models. For bidirectional (i.e., forward and backward) data processing, the Bi-LSTM model employs two hidden LSTM layers [32,33]. To do this, the main recursive layer of the neural model is duplicated. During training, the information that goes into the main layer is the real data, but the information that goes into the duplicate layer is the exact opposite. This strategy significantly enhances the quantity of data accessible to the model. The inner workings of a Bi-LSTM model. The Bi-LSTM forward hidden layer.

The Bi-LSTM forward hidden layer $(\vec{\beta})$, the backward hidden layer $(\overleftarrow{\beta})$, and the output $(o)$ can be deduced from the following equations [34].

\begin{equation}
\overrightarrow{\beta_t} = f(W^{\rightarrow}_{\beta} x_t + U^{\rightarrow}_{\beta}\overrightarrow{\beta}_{t-1} + b^{\rightarrow})
\end{equation}

\begin{equation}
\overleftarrow{\beta_t} = f(W^{\leftarrow}_{\beta} x_t + U^{\leftarrow}_{\beta}\overleftarrow{\beta}_{t+1} + b^{\leftarrow})
\end{equation}

\begin{equation}
o_t = g(V_{\beta}[\overrightarrow{\beta_t},\overleftarrow{\beta_t}] + c)
\end{equation}

In this context, $\hat{\mathcal{W}}^{\alpha}_{\beta \rightarrow}$ represents the backward hidden weight and $\hat{\mathcal{W}}^{\alpha}_{\rightarrow \beta}$ represents the forward hidden weight. The term $\mathfrak{h}$ refers to the hidden layer, whereas $\lambda_{\beta \rightarrow}$ and $\lambda_{\rightarrow \beta}$ denote the forward and backward bias vectors, respectively [35].

Gated Recurrent Unit (GRU) was created to tackle the issue of gradients that disap- pear or explode. It is an enhanced version of the LSTM model that also uses gate struc- tures to regulate the flow of information. The absence of an output gate at GRU means that all data is accessible to the whole public, which is a significant observation [36]. GRUs only have two gates—one for updating and one for resetting—but LSTMs com- bine input and forget gates. The simplified structure of GRUs allows them to increase performance with fewer parameters. Following equations describe the GRU update and reset gates:

\begin{equation}
m_t = \sigma \left( W_m [h_{t-1}, x_t] + U_m h_{t-1} + b_m \right)
\end{equation}

\begin{equation}
n_t = \sigma \left( W_n [h_{t-1}, x_t] + U_n h_{t-1} + b_n \right)
\end{equation}

where $m_t$ is the reset gate at time step $t$, $n_t$ is the update gate at time step $t$, $h_{t-1}$ is the hidden state at time step $t-1$, $x_t$ is the input at time step $t$, $W_m$ and $W_n$ are the weight matrices for the reset and update gates, $U_m$ and $U_n$ are the weight matrices applied to the hidden state, $b_m$ and $b_n$ are the bias vectors for the reset and update gates, and $\sigma$ denotes the sigmoid activation function. The candidate hidden state $\tilde{h}_t$ is then computed as follows.

\begin{equation}
\tilde{h}_t = \tanh \left( W [m_t \ast h_{t-1}, x_t] + b \right)
\end{equation}

The position of the carry gate $(1 - n_t)$ determines the extent to which the previous hidden state is carried forward, as well as the range of values through which the candidate hidden state is used to update the previous hidden state.

\begin{equation}
h_t = (1 - n_t) h_{t-1} + n_t \tilde{h}_t
\end{equation}

\subsection{Performance Metrics}

The evaluation model consists of four components: TP, TN, FP, and FN. 
The term \textit{True Positive} (TP) refers to the correct identification of a positive sample. 
Similarly, \textit{True Negative} (TN) represents the correct prediction of a negative sample. 
When a positive sample is incorrectly predicted as negative, it is referred to as a \textit{False Negative} (FN), 
whereas a \textit{False Positive} (FP) occurs when a negative sample is incorrectly predicted as positive.

The performance of the model is evaluated using Equations~(11)--(14), where 
$\mathrm{Accuracy}$, $\mathrm{F1\text{-}score}$, $\mathrm{Precision}$, and $\mathrm{Recall}$ 
are the performance metrics. Accuracy is defined as the ratio of correctly predicted samples 
to the total number of predictions made by the model. The $\mathrm{F1\text{-}score}$, also known 
as the harmonic mean of $\mathrm{Precision}$ and $\mathrm{Recall}$, is used to measure the overall 
performance of the model. Recall measures the proportion of actual positive samples that are 
correctly identified by the model. Models with values closer to 1 generally indicate better 
performance according to the ROC curve.

\begin{equation}
Accuracy = \frac{TP + TN}{TP + TN + FP + FN}
\end{equation}

\begin{equation}
Precision = \frac{TP}{TP + FP}
\end{equation}

\begin{equation}
Recall = \frac{TP}{TP + FN}
\end{equation}

\begin{equation}
F1\text{-}score = \frac{2 \times Precision \times Recall}{Precision + Recall}
\end{equation}

\section{Results and Analysis}

\subsection{Results of Bi-LSTM}

The plot of Training Loss and Validation Loss over a series of epochs during the train- ing of an ML model is shown in Fig. 2. The number of iterations or training cycles the model has gone through from 0 to around 30 epochs. The loss value measures how well the model is performing. A lower loss indicates a better model performance. The model initially starts with a higher loss (around 0.15) at epoch 1, but within the first few epochs (about 5), the training loss decreases sharply, indicating rapid learning. After epoch 5, the training loss continues to decrease but at a much slower rate, stabilizing at around 0.055. Like the training loss, the validation loss also starts high, around 0.1, and quickly drops, nearly matching the training loss around epoch 5. After that, it follows a similar pattern, stabilizing slightly above the training loss (around 0.06). Both the training and validation losses are decreasing, indicating that the model is learning effectively without significant over-fitting. 

\begin{figure}[H]
\centering
\includegraphics[width=0.4\textwidth]{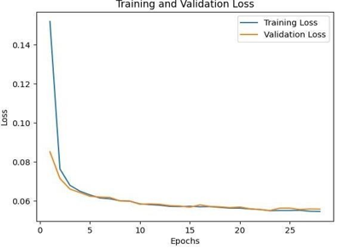}
\caption{Training and validation loss of the Bi-LSTM model across epochs}
\label{fig:bilstm_loss}
\end{figure}

The training and validation accuracy of a Bi-LSTM model across 30 epochs is shown in Fig. 3. The accuracy of the model is a measure of the proportion of correctly classi- fied instances.

\begin{figure}[H]
\centering
\includegraphics[width=0.3\textwidth]{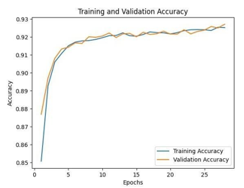}
\caption{Training and validation accuracy of the Bi-LSTM model across epochs}
\label{fig:bilstm_accuracy}
\end{figure}

\begin{figure}[H]
\centering
\includegraphics[width=0.3\textwidth]{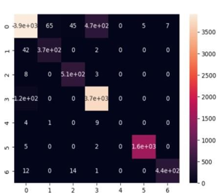}
\caption{Confusion matrix of the Bi-LSTM model}
\label{fig:bilstm_confusion}
\end{figure}

The accuracy starts from around 85\% (0.85) and rises to just above 92\% (0.92). It starts at approximately 85\% accuracy at epoch 1 and rises steeply in the first 5 epochs, reach- ing above 90\%. After that, the training accuracy gradually increases and stabilizes around 92\% to 93\%, showing minimal fluctuation. It closely follows the training accu- racy, starting slightly below the training accuracy but converging to a similar range between 92\% and 93\% after around 5 epochs. There are some slight fluctuations in validation accuracy, but overall, it remains close to the training accuracy.

Table~2 presents the evaluation results of the Bi-LSTM model, achieving an $\mathrm{Accuracy}$ of $92.7\%$, $\mathrm{Precision}$ of $92.9\%$, $\mathrm{Recall}$ of $92.7\%$, and an $\mathrm{F1\text{-}score}$ of $92.7\%$.

\begin{table}[H]
\centering
\caption{Evaluation of Bi-LSTM}
\label{tab:bilstm_eval}
\begin{tabular}{|l|c|}
\hline
\textbf{Parameters} & \textbf{Value (\%)} \\
\hline
Accuracy & 92.7 \\
\hline
Precision & 92.9 \\
\hline
Recall & 92.7 \\
\hline
F1-score & 92.7 \\
\hline
\end{tabular}
\end{table}

\subsection{Results of GRU Model}

Fig. 5 illustrates the training and validation loss over 25 epochs. Both losses start high, around 0.18, and decrease rapidly within the first few epochs, indicating that the model is learning effectively. By epoch 5, both training and validation losses converge near 0.06. After this point, the losses continue to gradually decrease and stabilize around 0.04 until epoch 20. Around epoch 20, there is a slight increase in both training and validation losses, which could signal slight overfitting or model instability, but the losses remain low overall. The model seems to generalize well, as the training and validation losses follow a similar trend and are closely aligned.

\begin{figure}[H]
\centering
\includegraphics[width=0.4\textwidth]{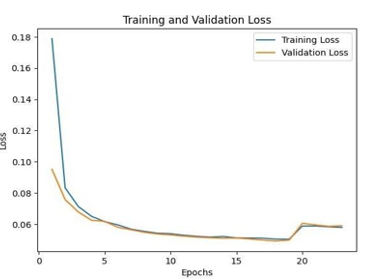}
\caption{Training and validation loss of the GRU model across epochs}
\label{fig:gru_loss}
\end{figure}

Fig. 6 shows the training and validation accuracy curves over 25 epochs for a GRU model. Both the training and validation accuracies start around 0.82 and improve significantly within the first 5 epochs. The training accuracy rises rapidly and stabilizes around 0.93, while the validation accuracy follows a similar trend but with more fluctuations. Around epoch 10, both accuracies converge near their peak (above 0.92), indicating good model performance. However, around epoch 20, there is a sudden drop in validation accuracy, possibly suggesting some over-fitting or instability in the model's generalization capability at this point. After the drop, the validation accuracy recovers slightly but remains below the earlier peak. Overall, the model performs well, with both accuracies stabilizing after initial training. Fig. 7 shows the confusion matrix of GRU model.

\begin{figure}[H]
\centering
\includegraphics[width=0.4\textwidth]{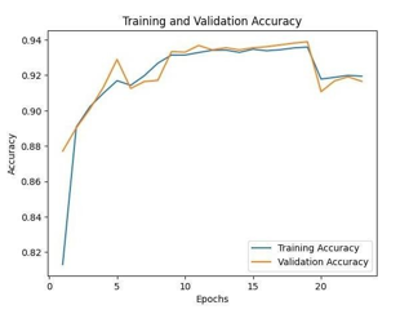}
\caption{Training and validation accuracy of the GRU model across epochs}
\label{fig:gru_accuracy}
\end{figure}

\begin{figure}[H]
\centering
\includegraphics[width=0.4\textwidth]{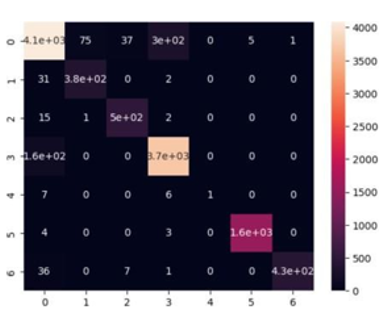}
\caption{Confusion matrix of the GRU model}
\label{fig:gru_confusion}
\end{figure}

Table~3 provides the evaluation results of the GRU model, which achieved an $\mathrm{Accuracy}$ of $93.9\%$, $\mathrm{Precision}$ of $94.0\%$, $\mathrm{Recall}$ of $93.9\%$, and an $\mathrm{F1\text{-}score}$ of $93.8\%$.

\begin{table}[H]
\centering
\caption{Evaluation of GRU}
\label{tab:gru_eval}
\begin{tabular}{|l|c|}
\hline
\textbf{Parameters} & \textbf{Value (\%)} \\
\hline
Accuracy & 93.9 \\
\hline
Precision & 94 \\
\hline
Recall & 93.9 \\
\hline
F1-score & 93.8 \\
\hline
\end{tabular}
\end{table}

\subsection{Results of Hybrid Model}

Fig. 8 illustrates the training and validation accuracy of a hybrid model (Bi-LSTM- GRU) over 25 epochs.

\begin{figure}[H]
\centering
\includegraphics[width=0.4\textwidth]{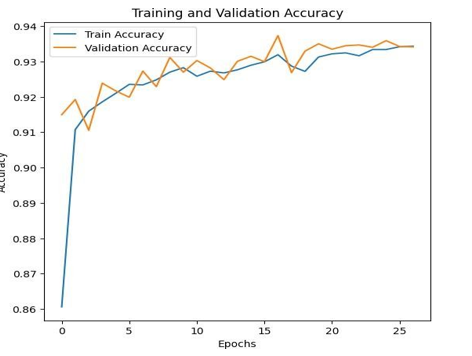}
\caption{Accuracy of Hybrid Model}
\label{fig:hybrid_loss}
\end{figure}

At the beginning (Epoch 1), the training accuracy is slightly above 0.86. It rapidly improves, reaching around 0.90 at the 5th epoch. The accuracy then continues to gradually increase, stabilizing around 0.93 toward the 20th epoch, with minor fluctuations. The final training accuracy at epoch 25 is slightly below 0.93. Starting at a value similar to the training accuracy (~0.88), the validation accuracy also increases, but its progress is less steady. It shows more fluctuations compared to the training accuracy, with noticeable peaks and dips. Around epoch 15, a sharp increase is observed, followed by a drop and subsequent stabilization. The final validation accuracy is about 0.93, matching the training accuracy.

\begin{figure}[H]
\centering
\includegraphics[width=0.4\textwidth]{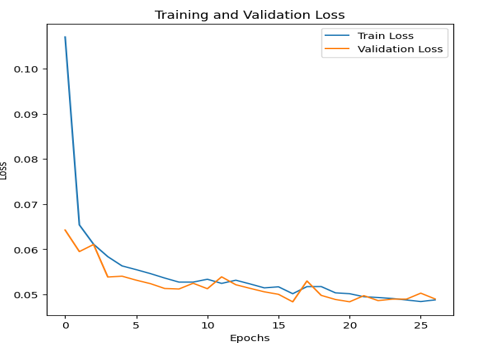}
\caption{Loss of Hybrid Model}
\label{fig:hybrid_accuracy}
\end{figure}

\begin{figure}[H]
\centering
\includegraphics[width=0.4\textwidth]{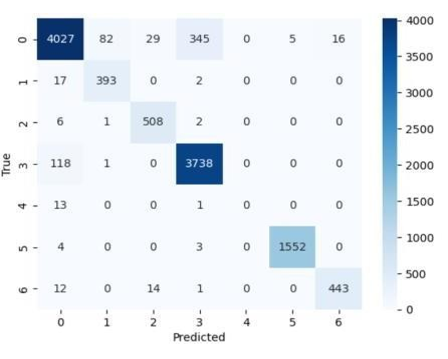}
\caption{Confusion Matrix of Hybrid Model}
\label{fig:hybrid_accuracy}
\end{figure}

Figure 9 shows the loss during training and validation of a hybrid model over an interval of 25 epochs. The training loss is quite significant at around 0.11 at epoch 1, indicating that the model's predictions are detached from the predicted values. It rapidly decreases in the first few epochs, reaching around 0.07 by epoch 5. The loss continues to decline gradually as the epochs progress, stabilizing at around 0.03-0.04 after about 15 epochs. The validation loss also begins relatively high, similar to the training loss, but it quickly decreases and follows a similar trend. By the 5th epoch, it stabilizes around 0.05, with slight fluctuations but staying below the training loss in many parts of the plot. The overall stability in the validation loss suggests that the model is not over fitting and is generalizing well to unseen data. Both training and validation loss values converge and decrease steadily, representative that the model is learning effectively over time. Figure 10 show the confusion matrix of hybrid model.

\begin{table}[H]
\centering
\caption{Evaluation of the Hybrid Model}
\label{tab:hybrid_eval}
\begin{tabular}{|l|c|}
\hline
\textbf{Parameters} & \textbf{Value (\%)} \\
\hline
Accuracy & 97.07 \\
\hline
Precision & 94.13 \\
\hline
Recall & 94.07 \\
\hline
F1-score & 94 \\
\hline
\end{tabular}
\end{table}

Table~4 provides the evaluation results of the proposed hybrid model, which achieved an $\mathrm{Accuracy}$ of $97.07\%$, $\mathrm{Precision}$ of $94.13\%$, $\mathrm{Recall}$ of $94.07\%$, and an $\mathrm{F1\text{-}score}$ of $94\%$.

\subsection{Comparative Analysis}

The study demonstrates significant advancements in CAD detection, with the hybrid Bi-LSTM+GRU model achieving higher accuracy, precision, recall, and F1 scores compared to existing methods which is demonstrated in Table 5.

\begin{table}[H]
\centering
\caption{Evaluation of existing models compared to the proposed model}
\label{tab:model_comparison}
\begin{tabular}{|l|l|c|}
\hline
\textbf{Author (Year) [Reference]} & \textbf{Methodology} & \textbf{Accuracy (\%)} \\
\hline
Tang et al., (2023) [38] & CNN & 70 \\
\hline
Sapra et al., (2023) [39] & DNN & 95.2 \\
\hline
Choi et al., (2023) [40] & Ob-CAD & 63.8 \\
\hline
Our Work & Hybrid Model & 97.07 \\
\hline
\end{tabular}
\end{table}

The proposed model performs better than previous approaches, achieving an $\mathrm{Accuracy}$ of $94.07\%$, $\mathrm{Precision}$ of $94.13\%$, $\mathrm{Recall}$ of $94.07\%$, and an $\mathrm{F1\text{-}score}$ of $94\%$. The graphical representation of the comparison results is illustrated in Fig 11.

\begin{figure}[H]
\centering
\includegraphics[width=0.4\textwidth]{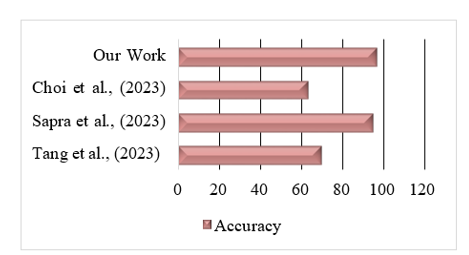}
\caption{Comparison of previous work v/s present work}
\label{fig:hybrid_accuracy}
\end{figure}

The integration of non-invasive data sources reduces patient risks and costs, enabling timely and accurate diagnosis. This approach has the potential to significantly improve clinical decision-making, reducing dependency on invasive diagnostics and accelerating personalized treatment strategies. Machine learning revolutionizes CAD diagnosis by enabling robust analysis of large and complex datasets, identifying patterns and risk factors that traditional methods might miss. Advanced ML models like Bi-LSTM and GRU process temporal and sequential data effectively, making them ideal for analyzing clinical trends and imaging data. The scalability and adaptability of ML techniques make them indispensable for early detection and monitoring of CAD, fostering preventive care, and improving patient outcomes globally.

\section{Conclusion}

CAD is one of the leading causes of mortality worldwide, necessitating improved diagnostic techniques. Traditional diagnostic methods, such as angiography and stress tests, are often invasive, expensive, or time-consuming. In recent years, the application of ML has emerged as a promising approach for enhancing CAD detection. This paper investigates the various applications of ML algorithms and techniques that can improve the accuracy and efficiency of CAD diagnosis. Machine learning models can identify patterns indicative of CAD with higher precision by analyzing patient data, including clinical records, imaging, and biomarkers. Key models, including Bi-LSTM, GRU, and hybrid models, are used for the detection of CAD on the 3D CCTA dataset. The experimental results show that the hybrid model achieved an accuracy of 97.07\%. The findings suggest that machine learning can significantly enhance CAD detection, leading to better patient outcomes through timely intervention and personalized treatment strategies.

\nocite{*}

\bibliographystyle{plain}
\bibliography{references}
\end{document}